\documentclass{article}

\usepackage[nonatbib,preprint]{neurips_data_2021}

\usepackage[utf8]{inputenc} % allow utf-8 input
\usepackage[T1]{fontenc}    % use 8-bit T1 fonts
\usepackage{hyperref}       % hyperlinks
\usepackage{url}            % simple URL typesetting
\usepackage{booktabs}       % professional-quality tables
\usepackage{amsfonts}       % blackboard math symbols
\usepackage{nicefrac}       % compact symbols for 1/2, etc.
\usepackage{microtype}      % microtypography
\usepackage{xcolor}         % colors
\usepackage{amsmath}
\usepackage{multirow}
\usepackage{graphicx}
\usepackage{amsthm}
\usepackage{booktabs}
\usepackage{amssymb}
\usepackage{subfigure}

\usepackage{multirow}
\usepackage{ulem}

\usepackage{makecell}

\usepackage{algorithm,algpseudocode}
\usepackage{amssymb}
\usepackage{booktabs}

\usepackage{caption}
\usepackage{graphicx}

\usepackage{nameref}

\title{\textit{ASR-GLUE}: A New Multi-task Benchmark for \\ ASR-Robust Natural Language Understanding}

\author{%
    Lingyun Feng\\
    Tsinghua University\\
    Tencent AI Lab\\
    Shen Zhen, China\\
    \texttt{fly19@mail.tsinghua.edu.cn}\\
    \And
    Jianwei Yu\\
    Tencent AI Lab\\
    Shen Zhen, China\\
    \texttt{tomasyu@tencent.com}\\
    \And
    Deng Cai\\
    The Chinese University of Hong Kong\\
    Hong Kong, China\\
    \texttt{thisisjcykcd@gmail.com}\\
    \And
    Songxiang Liu\\
    Tencent AI Lab\\
    Shen Zhen, China\\
    \texttt{shaunxliu@tencent.com}\\
    \And
    Haitao Zheng\\
    Tsinghua University\\
    Shen Zhen, China\\
    \texttt{zheng.haitao@sz.tsinghua.edu.cn}\\
    \And
    Yan Wang\\
    Tencent AI Lab\\
    Shen Zhen, China\\
    \texttt{brandenwang@tencent.com}\\
}  
\begin{document}

\maketitle

\begin{abstract}
Language understanding in speech-based systems has attracted attention in recent years with the growing demand for voice interface applications.
However, the robustness of natural language understanding (NLU) systems to errors introduced by automatic speech recognition (ASR) is under-examined. 
In this paper, we propose \textbf{\textit{ASR-GLUE}} benchmark, a new collection of 6 different NLU tasks for evaluating the performance of models under ASR error across 3 different levels of background noise and 6 speakers with various voice characteristics. 
Based on the proposed benchmark, we systematically investigate the effect of ASR error on NLU tasks in terms of noise intensity, error type and speaker variants. The analysis of this dataset shows that NLU under ASR errors is still very challenging and requires further research.
\footnote{The dataset is available at \\ \url{https://drive.google.com/drive/folders/1slqI6pUiab470vCxQBZemQZN-a_ssv1Q}}
\end{abstract}

\section{Introduction}

Language understanding in speech-based systems has attracted much attention in recent years with the growing demand for voice interface applications and devices such as Alexa~\cite{wang2020data}, Siri~\cite{williams2007partially}, and Cortana~\cite{wang2018modelling}. These speech-based intelligent systems usually comprise an automatic speech recognition (ASR) component which converts audio signals to readable natural language text, and a natural language understanding (NLU) component which takes the output of the ASR component as input and fulfills downstream tasks such as sentiment analysis, natural language inference, and response selection.
The upstream ASR error may propagate to the downstream NLU component and degrade the overall performance~\cite{serdyuk2018towards,wang2020large}. In real-world scenarios, ASR error can be ubiquitous due to poor articulation and acoustic variability caused by environment noise and reverberation~\cite{errattahi2018automatic}. The persistence of ASR error indicates a need for ASR-robust natural language understanding.

Previous work in this area is limited to task-oriented language understanding such as hotel reservation and meeting scheduling through human-machine interactions ~\cite{schumann2018incorporating,weng2020joint,rao2020speech,huang2020learning}. However, ASR error can affect many other NLU tasks, such as sentiment analysis in voice assistants. A benchmark that enables the comprehensive evaluation of NLU under ASR error on a diverse range of tasks is still missing.

In this paper, to quantitatively investigate how ASR error affects NLU capability, we propose the ASR-robust General Language Understanding Evaluation (\textbf{\textit{ASR-GLUE}}) benchmark: a collection of 6 NLU tasks including sentiment analysis, similarity and paraphrase tasks, and natural language inference (NLI) tasks. We hire 6 native speakers to convert the test data into audio recordings with 3 different levels of environment noise. Each speaker is requested to record all test data to study the variance between individuals.

Finally, we get 18 different types of audio recordings (3 levels of noise * 6 different speakers) for each of the 6 NLU tasks, varying in noise intensity, error type, and speaker variants. In addition, we also test human performance under different noise levels.  We hope it would benefit the research of ASR-robust NLU in the future.

Our contributions are as follows: 1) A new benchmark dataset, \textbf{\textit{ASR-GLUE}}, is proposed 
to enable a comprehensive evaluation of the robustness of NLU model to ASR error, covering 6 diversified tasks under 6 different speakers and 3 different levels of environment noise. 
2) We systematically and quantitatively investigate the sensitivity of  state-of-the-art NLU models to ASR error in terms of  noise intensity, error type and speaker variants.

\section{\textbf{\textit{ASR-GLUE}}}
\label{sec:data}

\subsection{Selected NLU Tasks}\label{Selected Datasets}

 The proposed \textbf{\textit{ASR-GLUE}} is constructed on the basis of GLUE~\cite{wang2018glue}, a popular NLU evaluation benchmark consisting of diverse NLU tasks. We select 5 typical NLU tasks from it, namely: Sentiment classification (SST-2~\cite{socher2013recursive}), Semantic Textual Similarity (STS-B~\cite{cer2017semeval}), paraphrase (QQP~\footnote{\url{data.quora.com/First-Quora-Dataset-Release-Question-Pairs}}), Question-answering NLI (QNLI~\cite{rajpurkar2016squad}), Recognizing Textual Entailment (RTE~\cite{dagan2005pascal,haim2006second,giampiccolo2007third,bentivogli2009fifth}.) and incorporate with a Science NLI task (SciTail~\cite{khot2018scitail}), resulting in 6 tasks in total. They are common and typical tasks of language understanding in speech-based scenarios, making them suitable for \textbf{\textit{ASR-GLUE}}. These tasks are described in detail below.

\textbf{SST-2} The Stanford Sentiment Treebank~\cite{socher2013recursive} is a single-input understanding task for sentiment classification. The task is to predict the sentiment of a given sentence in the movie reviews domain. Accuracy (ACC) of the binary classification (positive or negative) is used as the metric. 

\textbf{STS-B} The Semantic Textual Similarity Benchmark~\cite{cer2017semeval} consists of sentence pairs drawn from news headlines, video and image captions, and natural language inference data. The task is to predict sentence similarity scores which range from 1 to 5. We evaluate using Pearson and Spearman correlation coefficients.

\noindent\textbf{QQP} The Quora Question Pairs~\footnote{ data.quora.com/First-Quora-Dataset-Release-Question-Pairs} dataset consists of question pairs in social QA questions domain. The task is to determine whether a pair of questions are semantically equivalent. Accuracy (ACC) is used as the metric.

\noindent\textbf{QNLI} Question-answering NLI is modified from the Stanford Question Answering dataset~\cite{rajpurkar2016squad}. This is a sentence pair classification task that determines whether the context sentence contains the answer to the question. Accuracy (ACC) is used as the metric.

\noindent\textbf{SciTail} SciTail~\cite{khot2018scitail} is a recently released challenging textual entailment dataset collected from the science domain. This is a natural language inference task that determines if a natural language hypothesis can be justifiably inferred from a given premise. Accuracy (ACC) is used as the metric.

\noindent\textbf{RTE} The Recognizing Textual Entailment (RTE) datasets come from a series of annual textual

entailment challenges merged from a collection of ~\cite{dagan2005pascal,haim2006second}. All datasets are combined and converted to two-class classification: entailment and not entailment. Accuracy (ACC) is used as the metric.

\subsection{Data Construction}
Since the original datasets are presented in clean text form, we manually select instances from their test sets for human audio recording to evaluate the NLU capability in the presence of ASR error. Samples with non-standard words (e.g., abbreviations, currency expressions, strange names, and addresses) or sentences that are too long (more than 100 words) are excluded from these selected test sets. Considering the cost and quality of annotation, we keep the original training set and randomly select a subset of samples from the test set for human audio recording on each task\footnote{If there is no public test set, we use their dev set instead.}. The statistics of the data is shown in Table~\ref{tab:data}. 

\begin{figure*}[t] 
	\centering    
\includegraphics[width=0.99\textwidth]{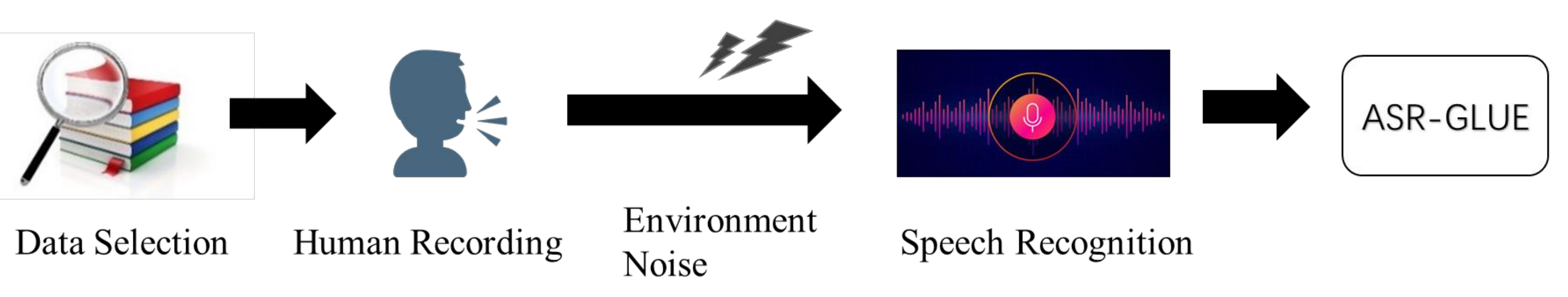}
	\caption{An illustration of the data collection and recording process.}
    \label{fig:overview}
\end{figure*}

In the human recording process, 6 native speakers are hired to record all test samples. Different levels of environment noise is provided and the audio signals are sent into an ASR system to get the final ASR hypothesis. The overall process is depicted in Figure~\ref{fig:overview}.

\begin{table}[H]
	\centering    
	\caption{Statistics of \textbf{\textit{ASR-GLUE}}. The reported hours are the sum of recording time under different levels of noise. We also report
	Word Error Rates (WER) for test sets under each noise level.}
\begin{tabular}{l|lll|ccc|c}
\hline
\multirow{2}{*}{Corpus} & \multirow{2}{*}{\#Train} & \multirow{2}{*}{\#Dev} & \multirow{2}{*}{\#Test} & \multicolumn{3}{c|}{WER (test)}              & \multirow{1}{*}{Hours} \\
\cline{5-8}
                        &                              &                    &                          & Low Noise & Medium Noise & High Noise &  Test+Dev                      \\ \hline
SST-2                   & 67349                    & 2772                   & 2790                    & 18.35\%   & 30.76\%      & 34.86\%    & 8.05                   \\
STS-B                   & 5749                     & 3042                   & 3222                    & 12.49\%   & 24.70\%      & 28.53\%    & 10.82                  \\
QQP                     & 363846                   & 1476                   & 3996                    & 13.78\%   & 24.34\%      & 27.45\%    & 11.56                  \\
QNLI                    & 104743                   & 2718                   & 2718                    & 22.51\%   & 33.61\%      & 37.90\%    & 18.00                  \\
RTE                     & 2490                     & 2070                   & 2088                    & 24.54\%   & 39.47\%      & 47.03\%    & 26.19                  \\
SciTail                 & 23596                    & 2718                   & 2736                    & 17.55\%   & 30.09 \%     & 34.04\%    & 16.81                  \\ \hline
\end{tabular}
\label{tab:data}
\end{table}

\paragraph{Human Audio Recording}\label{Recording}
We hire six native speakers to record the test sets for each task. The speakers are 3 males and 3 females with different ages ranging from 18 to 40 from the U.S. In the recording process, each speaker is required to record all text samples independently so that we can study the impact of speaker variation.  They are instructed to imagine they are communicating with someone when speaking the sentences of the six tasks. Note that they are allowed to make minor changes to the original text for natural and smooth expression such as change \textit{cannot} to \textit{can't}. To collect high-quality audio, all the original speech signals are recorded in a low-noise environment.

\paragraph{Environment Noise}\label{Noise}
\label{sec:nsm}

In real-world scenarios, human audio often recorded with environment noise and reverberation \cite{wolfel2009distant}.
The presence of the background interference will lead to substantial performance degradation of current ASR systems \cite{barker2018fifth} and further affect the downstream NLU systems \cite{henderson2014second}.
Therefore, to better evaluate the robustness of NLU models in the noisy acoustic environment, speech data with different levels of noise is further provided in the \textbf{\textit{ASR-GLUE}} corpus.

% Due to the large amount of different noise and reverberation conditions in real world, recording noisy and reverberant speech data in a wide variety of environments is not trivial.
% To this end
In \textbf{\textit{ASR-GLUE}}, the widely-adopted simulation approach \cite{ko2017study} is used to introduce different levels of noise and reverberation into the low-noise audio signals.Specifically, the background noise caused by such as phone ringing, alarm clocks and incoming vehicles are randomly sampled and added to the original recordings with a signal-to-noise-ratio (SNR) from 10dB to 15dB. Here SNR is defined as:
$\text{SNR}:=10\log\frac{||{\bf{s}}_{target}||^2}{||{\bf{s}}_{noise}||^2}$, 
where ${\bf s}_{target}$ and ${\bf s}_{noise}$ denote the acoustic signal of the clean speech and the noise respectively. $||\bf s||^2 = {\bf s}^T{\bf s}$ denotes the signal power. In addition, the room reverberation is also introduced by involving the recorded audio signals with the Room Impulse Responses (RIRs) \footnote{The noise and RIR files can be found at \url{http://www.openslr.org/resources/28/rirs\_noises.zip}} generated by the image-source method \cite{habets2006room}. The simulation process totally covers 843 kinds of different background noise and 417 types of different RIRs.

Finally, for each recorded human audio signal, we get three versions: (1) Low-level noise, same as the original audio. (2) Medium-level noise which introduces reverberation and 15dB SNR level background noise into the original audio. (3) High-level noise, which introduces reverberation and 10dB SNR level background noise into the original audio. 
Then we build a 6000h trained ASR system based on the widely-used open-source Kaldi toolkit\footnote{\url{https://github.com/kaldi-asr/kaldi}}~\cite{povey2011kaldi,povey2016purely} to transcribe these audio files into text. Table~\ref{tab:asr-wer} shows the WER of this system under different speakers and different noise levels.

\begin{table}[htb]
\centering
\caption{Detailed kaldi-based ASR WER on ASR-GLUE test set}
\scalebox{1}{
\begin{tabular}{l|c|cccccc}
\toprule
Corpus  & \multicolumn{1}{l}{Noisy level} & Speaker1 & Speaker2 & Speaker3 & Speaker4 & Speaker5 & Speaker6 \\
\hline
SST-2   & \multirow{6}{*}{Low}            & 11.50    & 18.53    & 30.03    & 19.94    & 13.96    & 38.36    \\
STS-B   &                                 & 6.87     & 11.89    & 17.76    & 12.62    & 7.67     & 13.62    \\
QQP     &                                 & 10.44    & 13.05    & 25.46    & 13.10    & 10.77    & 25.49    \\
QNLI    &                                 & 13.47    & 17.43    & 31.05    & 19.88    & 14.55    & 30.42    \\
RTE     &                                 & 23.04    & 18.98    & 34.34    & 22.05    & 15.60    & 37.77    \\
SciTail &                                 & 8.47     & 13.98    & 23.49    & 16.83    & 9.18     & 27.53    \\
\hline
\hline
SST-2   & \multirow{6}{*}{Medium}         & 21.64    & 29.68    & 40.12    & 34.60    & 23.23    & 53.26    \\
STS-B   &                                 & 20.06    & 28.16    & 31.11    & 21.79    & 16.49    & 29.65    \\
QQP     &                                 & 18.85    & 22.13    & 39.01    & 19.90    & 18.26    & 40.39    \\
QNLI    &                                 & 22.98    & 31.75    & 40.09    & 32.82    & 20.98    & 47.32    \\
RTE     &                                 & 37.21    & 34.39    & 47.25    & 35.63    & 25.54    & 59.38    \\
SciTail &                                 & 18.71    & 26.72    & 34.75    & 33.20    & 15.89    & 44.13    \\
\hline
\hline
SST-2   & \multirow{6}{*}{High}           & 25.98    & 33.90    & 42.82    & 37.77    & 27.27    & 55.31    \\
STS-B   &                                 & 24.66    & 32.61    & 33.22    & 25.09    & 18.83    & 33.71    \\
QQP     &                                 & 21.64    & 24.45    & 41.52    & 22.11    & 22.23    & 43.77    \\
QNLI    &                                 & 28.09    & 36.05    & 42.96    & 35.53    & 24.76    & 51.73    \\
RTE     &                                 & 44.12    & 41.20    & 54.56    & 50.70    & 33.33    & 64.91    \\
SciTail &                                 & 23.13    & 29.89    & 36.60    & 37.65    & 19.93    & 48.55   \\
\bottomrule
\end{tabular}
}
\label{tab:asr-wer}
\end{table}

% {\color{red}{Then we adopt a widely-used LF-MMI based ASR system\footnote{https://github.com/kaldi-asr/kaldi/blob/master/egs/swbd/s5c/local/chain/run\_tdnn\_lstm.sh} implemented with the Kaldi \cite{povey2011kaldi}~\cite{povey2016purely} toolkit\footnote{https://github.com/kaldi-asr/kaldi} to transcribe these audio to text.}}

\section{Analysis of \textbf{\textit{ASR-GLUE}}}
In this section we give extensive analyses on the proposed \textbf{\textit{ASR-GLUE}} dataset. In Section~\ref{humanPerformance}, we obtain human performance to measure
the ceiling performance on the test set, 
Then we analyse the performance of recent SOTA NLU models across different levels of environment noise in Section~\ref{NLUperformance}. Furthermore, in Section~\ref{errorType} we categorize ASR errors into four types, and systematically investigate the impact of different error types on NLU performance. Finally, we analyse the effect of voice variation from different speakers in Section~\ref{Individual}.

\subsection{Human Performance}
\label{humanPerformance}

To obtain human performance under environment noise, we hire native annotators to predict labels of each test sample in audio form. 

The annotators are hired from a third-party company. To guarantee the annotators fully understand these tasks, we first give the annotators a brief training on each task in \textbf{\textit{ASR-GLUE}}. Then we ask them to take an exam before starting annotation. Only annotators who pass the exam will be hired. Finally, we have 3 annotators to measure the ceiling performance for each task in \textbf{\textit{ASR-GLUE}}. Details about the exam and annotation process are presented below.

In the annotation process, the annotators are required to predict the labels of each test sample according to the corresponding audio signals. Note that for each test sample we have 18 audio signals (3 levels of noise * 6 speakers), which is a big burden for the annotators. So we randomly select one audio from the six speakers with each noise level and report the average performance of the annotators on all tasks.

The exam is set to guarantee the annotators fully understand each task. In the exam, we randomly select 50 samples from the original datasets which are in text form for each task. Annotators who achieve at least 90\% accuracy on these samples will be hired.

\subsection{Performance of Existing NLU Models}
\label{NLUperformance}
\begin{figure}[t] 
	\centering    
\includegraphics[width=0.99\textwidth]{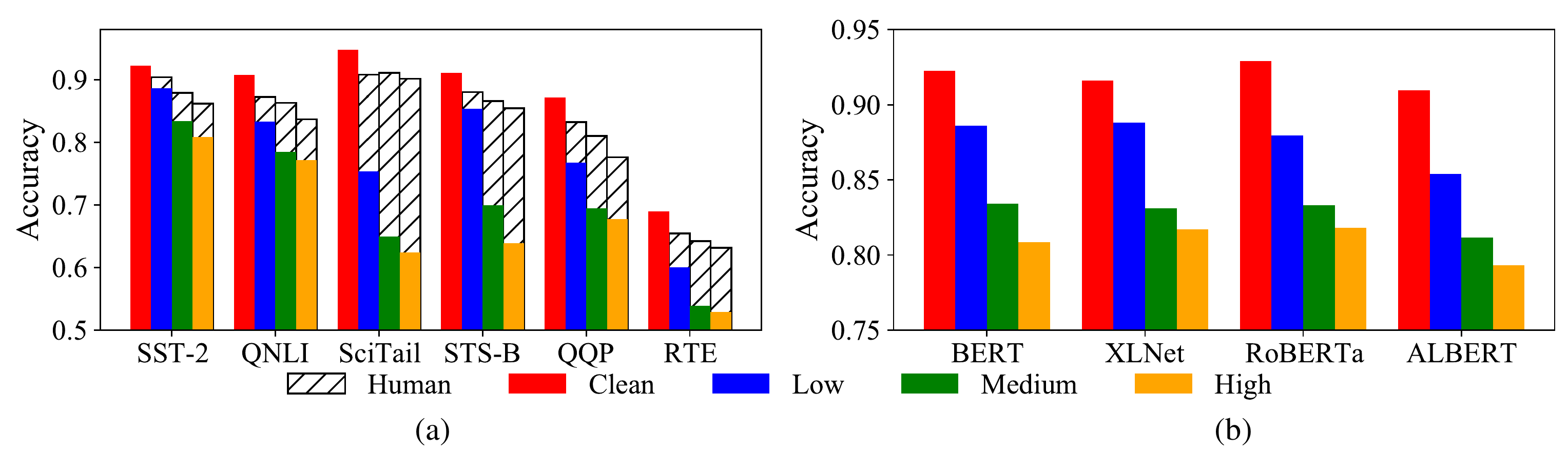}
\vspace{-0.2cm}
	\caption{(a) The performance of BERT on different tasks under different levels of noise. The shaded area represents human performance. (b) Accuracy results for different model architecture on SST-2 task. Here ``Human'' indicates human performance under various noise settings. ``Clean'' stands for test on clean text data. ``Low/Medium/High'' stands for test in low-level/medium-level/high-level noise respectively.
	}
    \label{fig:ex}
\end{figure}

To demonstrate the significance of the ASR error issue,  we leverage typical NLU models such as BERT to test their robustness to different levels of ASR error on different tasks. As shown in Figure~\ref{fig:ex}(a), While BERT yields promising results on error-free text, its performance degrades in the presence of ASR error on six tasks. As the noise increases, the performance of the model drops more severely. In contrast, humans are less affected by the environment noise, which indicates that there still remains a gap between the robustness of models and humans to ASR error.
% This indicates that there still remains a gap between the robustness of models and humans to environment noise. We can also observe that BERT is more vulnerable to ASR error on SciTail task while has relatively small degradation on SST-2 and QNLI tasks in the presence of ASR error.

We also investigate the effect of ASR error with different noise levels on different NLU models. We adopt base version of BERT~\cite{devlin2018bert}, RoBERTa~\cite{liu2019roberta}, ALBERT~\cite{lan2019albert}, XLNet~\cite{yang2019xlnet} as the NLU model. As shown in Fig.~\ref{fig:ex}(b) we can observe that
all these pretrained language models are sensitive to ASR error and the performance degrades with the increase of the noise level.

\subsection{Breakdown Analysis by ASR Error Type}
\label{errorType}
\begin{table}[]
\caption{Examples of speech recognition perturbation in \textbf{\textit{ASR-GLUE}}.}
\scalebox{0.87}{
\begin{tabular}{l|l|l}
\hline
Error Type           & Ground Truth              & Recognition Result                     \\ \hline
% Similar sounds & \makecell[l]{The {\bf{city}}  grew as an important centre for\\ the wool trade}                & \makecell[l]{The {\color{red}{silly}}  grew as an important centre for\\the wool trade}\\

\multirow{2}{*}{Similar sounds} & The man couldn't {\bf{lift}} his son. & The man couldn't {\color{red}{\bf{lived}}} his son. \\
&Tommy {\bf{dropped}} his ice cream. &  Tommy {\color{red}{\bf{jumped}}} his ice cream.\\
\hline
\multirow{2}{*}{Liaison}        & Does Quora {\bf{stand for}} question or answer. & Does Quora {\color{red}{\bf{Stanford}}} question or answer.               \\
&The {\bf{drain is}} clogged with hair. & The {\color{red}{\bf{drains}}} clogged with hair. \\
% \hline
% % Spoken numbers & 13:45              & thirteen forty five       \\
% % Pauses         & I want to go home. & I want to um go uh home.  \\
\hline
\multirow{2}{*}{Insertion}        & This afternoon.  & This {\color{red}{\bf after}} afternoon.    \\
& A warm funny $\ast$ engaging film & A warm funny {\color{red}{\bf{and}}} engaging film. \\
\hline
\multirow{2}{*}{Deletion}         &  A  black and white photo of an old train station.                   & A  black {\color{red}{$\ast$}} white {\color{red}{$\ast$}} of {\color{red}{$\ast$}} train station    \\ 
& Old style bicycle {\bf{parked on}} floor& Old style bicycle {\color{red}{$\ast$}} {\color{red}{$\ast$}} floor \\
\hline
\end{tabular}
}
\label{tab:ex}
\end{table}

We conclude that the ASR error types can be categorized into four-folds, namely similar sounds, liaison, insertion, and deletion. Specifically,  (i) Similar sounds happen when the ASR system sometimes wrongly identifies one word as another with similar pronunciation. (ii) Liaison constantly occurs between successive words which have sound fusion across word boundaries. (iii) Insertion happens when the ASR system makes word redundancies. (iv) Deletion happens when there are word omissions in the ASR hypothesis. Examples of each error type are presented in Table~\ref{tab:ex}.

\begin{figure*}[t] 
	\centering    
\includegraphics[width=0.8\textwidth]{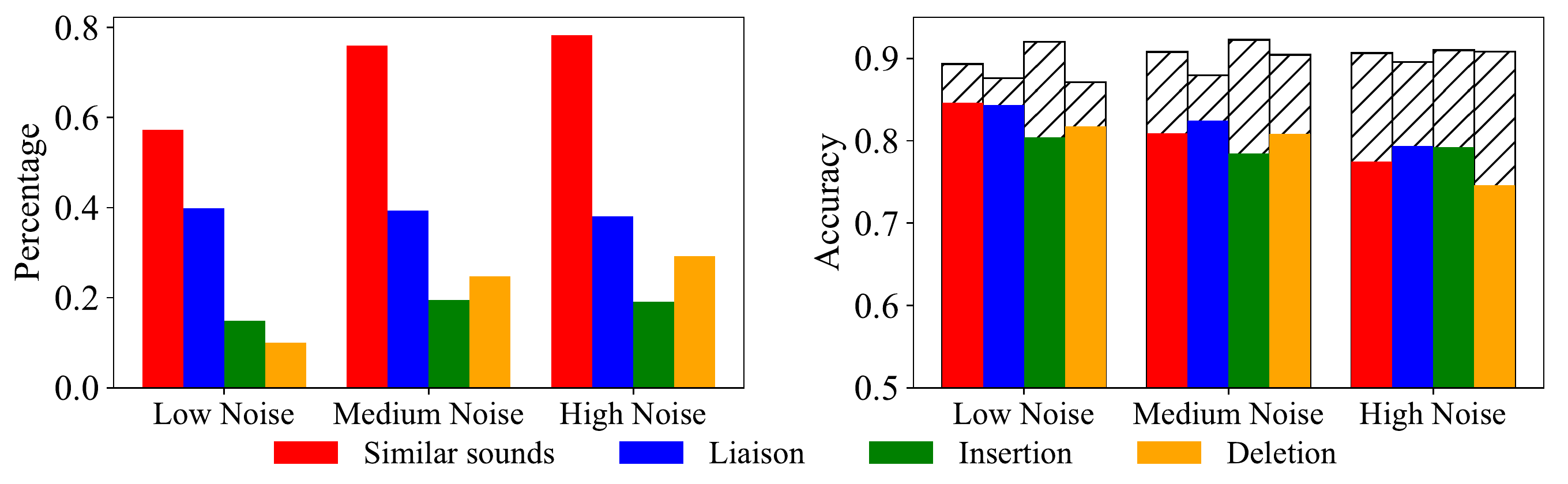}
\vspace{-0.5mm}
	\caption{Left: The percentage of each error type under different noise setting in SST-2 dataset. Right: The accuracy of BERT on four subsets under different noise level. Each subset only contains test samples with one specific error type. For example, the red block represents the accuracy of BERT on test samples which contain similar sounds errors. The shaded area represents the performance degradation against clean text caused by a specific error type.} 
    \label{fig:noise_type}
\end{figure*}

We report the percentage of each aforementioned error type in Figure~\ref{fig:noise_type}(a). We choose the SST-2 task as an example and observe that similar sounds most commonly happen. As the noise increases, the percentage of similar sounds and deletion made by the ASR system gradually increases while the percentage of liaison and insertion error types remain relatively stable. Note that the sum percentages of the four error types are over 100\%, since different error types may simultaneously exist in one hypothesis.  

% since ASR system can introduce more than one error type when transcribe a piece of voice data to text.

\label{sec:analysis}

We also investigate the impact of each error type on NLU model on SST-2 task. We group data according to error types and compare the model performance with the same grouped raw data without errors respectively. 
As shown in Figure~\ref{fig:noise_type} (b), we can observe that BERT can better handle Liaison error for the performance degradation is minimal under varied noise settings. As noise increases, the accuracy of the model decreases more severely for each different type of error.
%  similar sounds and deletion error have more impact on the NLU model, degrading the performance of the model more severely as the noise increase.
\begin{figure}[t] 
	\centering    
\includegraphics[width=0.99\textwidth]{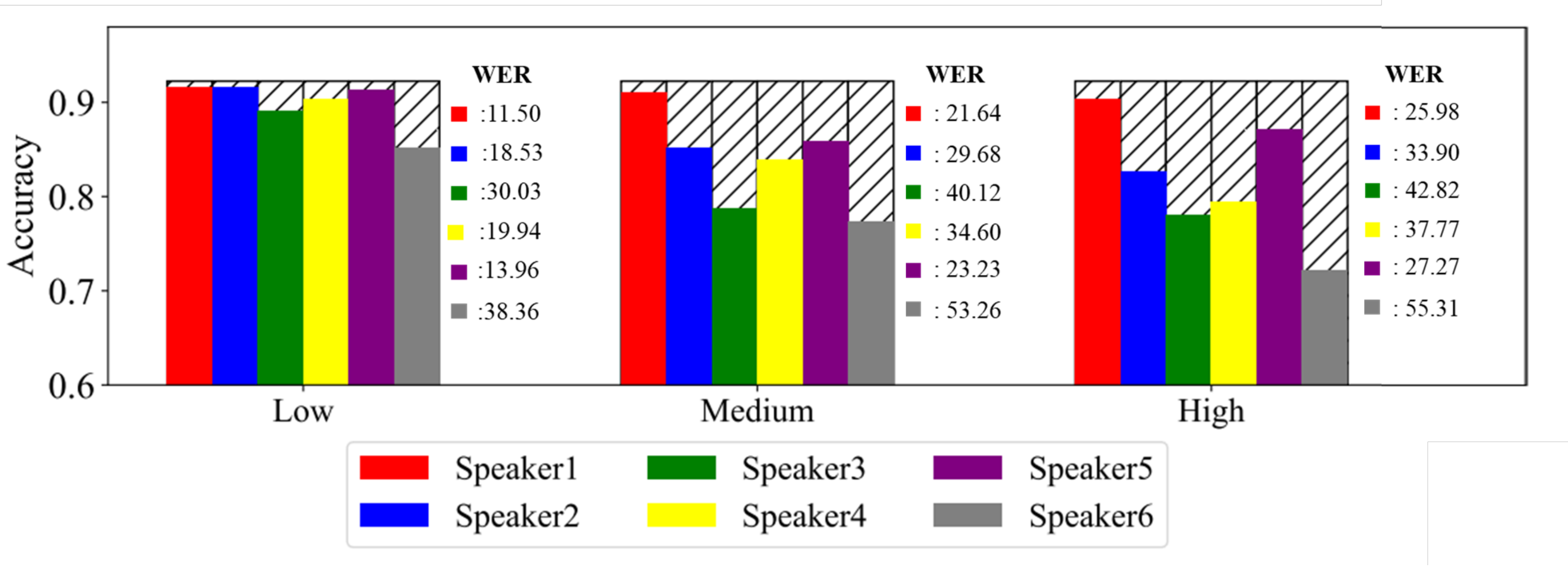}
\vspace{-0.2cm}
	\caption{NLU performance under voice variation from different speakers.}
    \label{fig:sp}
\end{figure}

\subsection{Effect of Individual Difference}
\label{Individual}

We test the effect of voice variation of the six hired speakers on the test set of SST-2. As shown in Fig.~\ref{fig:sp} we can observe that the recognition quality and classification accuracy vary greatly across speakers. For example, the accuracy of BERT on S1 (corresponds to Speaker 1) is consistently higher than S6 (corresponds to Speaker 6) with a very large margin ($27\% \sim 32\%$). Meanwhile, we can also observe that the drop in classification accuracy is due to the increase of WER. The higher WER means more ASR errors in the test samples, resulting in more misclassification. 

% ASR error distribution differs by speaker background variables, in turn affecting the downstream systems~\cite{zheng2005accent}.  
% We test the impact of different speakers on SciTail task. The six speakers differ by gender (3 male, 3 female), age and level of proficiency.
% As shown in Fig.~\ref{fig:sp} , we can observe that speaker 3, speaker 4, speaker 6 have more impact on ASR systems than speaker 1, speaker 2 and speaker 5 and then degrade the NLU performance more severely.

\section{Related Work}
Many benchmark datasets are  created to facilitate Spoken Language Understanding (SLU) ~\cite{coucke2018snips,lugosch2019speech,wang2020large,price1990evaluation,henderson2014second,peng2020raddle}, which evaluate the robustness of the downstream NLU model against the error output from the upstream acoustic model~\cite{schumann2018incorporating,weng2020joint,rao2020speech,huang2020learning}. 
However, they are only designed for a particular domain or a specific task such as intent detection and slot filling. In contrast, the human ability to understand language is general, flexible, and robust. There is a need to test general-purpose natural language understanding capability on diverse tasks in different domains.

Large-scale pretrained language models have achieved striking performance on NLU in recent years~\cite{jin2020bert,yang2019xlnet}. Recently many works test their robustness by human-crafted adversarial examples~\cite{nie2019adversarial} or generated examples by adversarial attacks~\cite{jin2020bert,madry2017towards,zhu2019freelb,dong2021towards}. 
~\cite{DBLP:conf/iclr/ZhaoDS18}  projects the input data to a latent space by generative adversarial networks (GANs),
and then retrieves adversaries close to the original instance in the latent space. ~\cite{iyyer2018adversarial} propose controlled paraphrase networks to generate syntactically adversarial examples that both fool pre-trained models and improve the robustness of these models to syntactic variation  when used to augment their training data.
However, the robustness of pre-trained model to speech recognition error in real conditions has not been fully explored.

% white-box attacks (Papernot et al., 2016; Li et al., 2019; Ebrahimi et al., 2018) based on the model’s internal structure or gradient signals.

% black-box attack: ~\cite{gao2018black} present a DeepWordBug algorithm to generate small perturbations in the character-level for black-box attack. They sort
% the tokens based on the importance evaluated by
% four functions, and make random token transformations such as substitution and deletion with the
% constraint of edit distance. 

\section{Conclusion}
We present \textbf{\textit{ASR-GLUE}}, a new benchmark for evaluating general-purpose language understanding under ASR error in speech-based applications. 
% The set of six tasks in
% our benchmark emphasizes diverse task formats and low-data training data tasks, with nearly half the
% tasks having fewer than 1k examples and all but one of the tasks having fewer than 10k examples.
We propose two ways to improve robustness of the NLU system and find that there is still a gap between the NLU capability of the model and humans.
\textbf{\textit{ASR-GLUE}} offers a rich and challenging testbed for work developing ASR robust model for general-purpose language understanding.
Given the difficulty of \textbf{\textit{ASR-GLUE}}, we expect that further progress in multi-task, multi-model learning techniques will be necessary to approach human-level performance on the benchmark.

\bibliography{neurips}

\begin{thebibliography}{10}

\bibitem{wang2020data}
Wang, L., M.~Fazel-Zarandi, A.~Tiwari, et~al.
\newblock Data augmentation for training dialog models robust to speech
  recognition errors.
\newblock \emph{arXiv preprint arXiv:2006.05635}, 2020.

\bibitem{williams2007partially}
Williams, J.~D., S.~Young.
\newblock Partially observable markov decision processes for spoken dialog
  systems.
\newblock \emph{Computer Speech \& Language}, 21(2):393--422, 2007.

\bibitem{wang2018modelling}
Wang, S., T.~Gunter, D.~VanDyke.
\newblock On modelling uncertainty in neural language generation for policy
  optimisation in voice-triggered dialog assistants.
\newblock In \emph{2nd Workshop on Conversational AI: Today’s Practice and
  Tomorrow’s Potential, NeurIPS}. 2018.

\bibitem{serdyuk2018towards}
Serdyuk, D., Y.~Wang, C.~Fuegen, et~al.
\newblock Towards end-to-end spoken language understanding.
\newblock In \emph{2018 IEEE International Conference on Acoustics, Speech and
  Signal Processing (ICASSP)}, pages 5754--5758. IEEE, 2018.

\bibitem{wang2020large}
Wang, P., L.~Wei, Y.~Cao, et~al.
\newblock Large-scale unsupervised pre-training for end-to-end spoken language
  understanding.
\newblock In \emph{ICASSP 2020-2020 IEEE International Conference on Acoustics,
  Speech and Signal Processing (ICASSP)}, pages 7999--8003. IEEE, 2020.

\bibitem{errattahi2018automatic}
Errattahi, R., A.~El~Hannani, H.~Ouahmane.
\newblock Automatic speech recognition errors detection and correction: A
  review.
\newblock \emph{Procedia Computer Science}, 128:32--37, 2018.

\bibitem{schumann2018incorporating}
Schumann, R., P.~Angkititrakul.
\newblock Incorporating asr errors with attention-based, jointly trained rnn
  for intent detection and slot filling.
\newblock In \emph{2018 IEEE International Conference on Acoustics, Speech and
  Signal Processing (ICASSP)}, pages 6059--6063. IEEE, 2018.

\bibitem{weng2020joint}
Weng, Y., S.~S. Miryala, C.~Khatri, et~al.
\newblock Joint contextual modeling for asr correction and language
  understanding.
\newblock In \emph{ICASSP 2020-2020 IEEE International Conference on Acoustics,
  Speech and Signal Processing (ICASSP)}, pages 6349--6353. IEEE, 2020.

\bibitem{rao2020speech}
Rao, M., A.~Raju, P.~Dheram, et~al.
\newblock Speech to semantics: Improve asr and nlu jointly via all-neural
  interfaces.
\newblock \emph{arXiv preprint arXiv:2008.06173}, 2020.

\bibitem{huang2020learning}
Huang, C.-W., Y.-N. Chen.
\newblock Learning asr-robust contextualized embeddings for spoken language
  understanding.
\newblock In \emph{ICASSP 2020-2020 IEEE International Conference on Acoustics,
  Speech and Signal Processing (ICASSP)}, pages 8009--8013. IEEE, 2020.

\bibitem{wang2018glue}
Wang, A., A.~Singh, J.~Michael, et~al.
\newblock Glue: A multi-task benchmark and analysis platform for natural
  language understanding.
\newblock \emph{arXiv preprint arXiv:1804.07461}, 2018.

\bibitem{socher2013recursive}
Socher, R., A.~Perelygin, J.~Wu, et~al.
\newblock Recursive deep models for semantic compositionality over a sentiment
  treebank.
\newblock In \emph{Proceedings of the 2013 conference on empirical methods in
  natural language processing}, pages 1631--1642. 2013.

\bibitem{cer2017semeval}
Cer, D., M.~Diab, E.~Agirre, et~al.
\newblock Semeval-2017 task 1: Semantic textual similarity-multilingual and
  cross-lingual focused evaluation.
\newblock \emph{arXiv preprint arXiv:1708.00055}, 2017.

\bibitem{rajpurkar2016squad}
Rajpurkar, P., J.~Zhang, K.~Lopyrev, et~al.
\newblock Squad: 100,000+ questions for machine comprehension of text.
\newblock \emph{arXiv preprint arXiv:1606.05250}, 2016.

\bibitem{dagan2005pascal}
Dagan, I., O.~Glickman, B.~Magnini.
\newblock The pascal recognising textual entailment challenge.
\newblock In \emph{Machine Learning Challenges Workshop}, pages 177--190.
  Springer, 2005.

\bibitem{haim2006second}
Haim, R.~B., I.~Dagan, B.~Dolan, et~al.
\newblock The second pascal recognising textual entailment challenge.
\newblock In \emph{Proceedings of the Second PASCAL Challenges Workshop on
  Recognising Textual Entailment}. 2006.

\bibitem{giampiccolo2007third}
Giampiccolo, D., B.~Magnini, I.~Dagan, et~al.
\newblock The third pascal recognizing textual entailment challenge.
\newblock In \emph{Proceedings of the ACL-PASCAL workshop on textual entailment
  and paraphrasing}, pages 1--9. 2007.

\bibitem{bentivogli2009fifth}
Bentivogli, L., P.~Clark, I.~Dagan, et~al.
\newblock The fifth pascal recognizing textual entailment challenge.
\newblock In \emph{TAC}. 2009.

\bibitem{khot2018scitail}
Khot, T., A.~Sabharwal, P.~Clark.
\newblock Scitail: A textual entailment dataset from science question
  answering.
\newblock In \emph{Proceedings of the AAAI Conference on Artificial
  Intelligence}, vol.~32. 2018.

\bibitem{wolfel2009distant}
W{\"o}lfel, M., J.~McDonough.
\newblock \emph{Distant speech recognition}.
\newblock John Wiley \& Sons, 2009.

\bibitem{barker2018fifth}
Barker, J., S.~Watanabe, E.~Vincent, et~al.
\newblock The fifth'chime'speech separation and recognition challenge: dataset,
  task and baselines.
\newblock \emph{arXiv preprint arXiv:1803.10609}, 2018.

\bibitem{henderson2014second}
Henderson, M., B.~Thomson, J.~D. Williams.
\newblock The second dialog state tracking challenge.
\newblock In \emph{Proceedings of the 15th annual meeting of the special
  interest group on discourse and dialogue (SIGDIAL)}, pages 263--272. 2014.

\bibitem{ko2017study}
Ko, T., V.~Peddinti, D.~Povey, et~al.
\newblock A study on data augmentation of reverberant speech for robust speech
  recognition.
\newblock In \emph{2017 IEEE International Conference on Acoustics, Speech and
  Signal Processing (ICASSP)}, pages 5220--5224. IEEE, 2017.

\bibitem{habets2006room}
Habets, E.~A.
\newblock Room impulse response generator.
\newblock \emph{Technische Universiteit Eindhoven, Tech. Rep}, 2(2.4):1, 2006.

\bibitem{povey2011kaldi}
Povey, D., A.~Ghoshal, G.~Boulianne, et~al.
\newblock The kaldi speech recognition toolkit.
\newblock In \emph{IEEE 2011 workshop on automatic speech recognition and
  understanding}, CONF. IEEE Signal Processing Society, 2011.

\bibitem{povey2016purely}
Povey, D., V.~Peddinti, D.~Galvez, et~al.
\newblock Purely sequence-trained neural networks for asr based on lattice-free
  mmi.
\newblock In \emph{Interspeech}, pages 2751--2755. 2016.

\bibitem{devlin2018bert}
Devlin, J., M.-W. Chang, K.~Lee, et~al.
\newblock Bert: Pre-training of deep bidirectional transformers for language
  understanding.
\newblock \emph{arXiv preprint arXiv:1810.04805}, 2018.

\bibitem{liu2019roberta}
Liu, Y., M.~Ott, N.~Goyal, et~al.
\newblock Roberta: A robustly optimized bert pretraining approach.
\newblock \emph{arXiv preprint arXiv:1907.11692}, 2019.

\bibitem{lan2019albert}
Lan, Z., M.~Chen, S.~Goodman, et~al.
\newblock Albert: A lite bert for self-supervised learning of language
  representations.
\newblock \emph{arXiv preprint arXiv:1909.11942}, 2019.

\bibitem{yang2019xlnet}
Yang, Z., Z.~Dai, Y.~Yang, et~al.
\newblock Xlnet: Generalized autoregressive pretraining for language
  understanding.
\newblock \emph{arXiv preprint arXiv:1906.08237}, 2019.

\bibitem{coucke2018snips}
Coucke, A., A.~Saade, A.~Ball, et~al.
\newblock Snips voice platform: an embedded spoken language understanding
  system for private-by-design voice interfaces.
\newblock \emph{arXiv preprint arXiv:1805.10190}, 2018.

\bibitem{lugosch2019speech}
Lugosch, L., M.~Ravanelli, P.~Ignoto, et~al.
\newblock Speech model pre-training for end-to-end spoken language
  understanding.
\newblock \emph{arXiv preprint arXiv:1904.03670}, 2019.

\bibitem{price1990evaluation}
Price, P.
\newblock Evaluation of spoken language systems: The atis domain.
\newblock In \emph{Speech and Natural Language: Proceedings of a Workshop Held
  at Hidden Valley, Pennsylvania, June 24-27, 1990}. 1990.

\bibitem{peng2020raddle}
Peng, B., C.~Li, Z.~Zhang, et~al.
\newblock Raddle: An evaluation benchmark and analysis platform for robust
  task-oriented dialog systems.
\newblock \emph{arXiv preprint arXiv:2012.14666}, 2020.

\bibitem{jin2020bert}
Jin, D., Z.~Jin, J.~T. Zhou, et~al.
\newblock Is bert really robust? a strong baseline for natural language attack
  on text classification and entailment.
\newblock In \emph{Proceedings of the AAAI conference on artificial
  intelligence}, vol.~34, pages 8018--8025. 2020.

\bibitem{nie2019adversarial}
Nie, Y., A.~Williams, E.~Dinan, et~al.
\newblock Adversarial nli: A new benchmark for natural language understanding.
\newblock \emph{arXiv preprint arXiv:1910.14599}, 2019.

\bibitem{madry2017towards}
Madry, A., A.~Makelov, L.~Schmidt, et~al.
\newblock Towards deep learning models resistant to adversarial attacks.
\newblock \emph{arXiv preprint arXiv:1706.06083}, 2017.

\bibitem{zhu2019freelb}
Zhu, C., Y.~Cheng, Z.~Gan, et~al.
\newblock Freelb: Enhanced adversarial training for natural language
  understanding.
\newblock \emph{arXiv preprint arXiv:1909.11764}, 2019.

\bibitem{dong2021towards}
Dong, X., A.~T. Luu, R.~Ji, et~al.
\newblock Towards robustness against natural language word substitutions.
\newblock In \emph{9th International Conference on Learning Representations
  (ICLR)}. 2021.

\bibitem{DBLP:conf/iclr/ZhaoDS18}
Zhao, Z., D.~Dua, S.~Singh.
\newblock Generating natural adversarial examples.
\newblock In \emph{6th International Conference on Learning Representations,
  {ICLR} 2018, Vancouver, BC, Canada, April 30 - May 3, 2018, Conference Track
  Proceedings}. OpenReview.net, 2018.

\bibitem{iyyer2018adversarial}
Iyyer, M., J.~Wieting, K.~Gimpel, et~al.
\newblock Adversarial example generation with syntactically controlled
  paraphrase networks.
\newblock \emph{arXiv preprint arXiv:1804.06059}, 2018.

\end{thebibliography}
\bibliographystyle{neurips}

\clearpage
\section{Datasheet}

\subsection{Dataset Motivation}
(1) For what purpose was the dataset created? Was there a specific task in mind? Was there a specific gap that needed to be filled? Please provide a description. 

This benchmark was created for the purpose of comprehensively investigate how ASR error affect NLU capability. We propose the ASR-robust General Language Understanding Evaluation (ASR-GLUE) benchmark: a new collection of 6 different NLU tasks for evaluating the performance of models under ASR error across 3 different levels of background noise and 6 speakers with various voice characteristics.

(2) Who created this dataset (e.g., which team, research group) and on behalf of which entity (e.g., company, institution, organization)?

Tencent AI Lab created and funded this dataset.

\subsection{Dataset Use Cases}
(1) What (other) tasks could the dataset be used for? 

The dataset can be used for sentiment analysis, paraphrase identification, natural language inference and many other NLU tasks. More importantly, this dataset can be used to comprehensively investigate how ASR error affect NLU capability.

(2) Are there tasks for which the dataset should not be used? If so, please provide a description 

No.

\subsection{Dataset Maintenance}
(1) Who is supporting/hosting/maintaining the dataset? How can the owner/curator/manager of the dataset be contacted? 

Tencent AI Lab will support and maintain the dataset. You can contact the owner via following e-mails:
Yan Wang, brandenwang@tencent.com
Jianwei Yu, tomasyu@tencent.com

(2) Is there an erratum? If so, please provide a link or other access point. 

No. 

(3) Will the dataset be updated? Will older versions of the dataset continue to be supported/hosted/maintained? 

We will update in future if necessary. We will support and maintain all versions.

(4) If others want to extend/augment/build on/contribute to the dataset, is there a mechanism for them to do so?

Others may do so. They just need notice the authors of this paper in advance.

\subsection{License}
% The authors of ASR-GLUE bear all responsibility in case of violation of rights, ethics of ASR-GLUE dataset itself.
The authors bear all responsibility in case of violation of rights.

% \doclicenseThis

% The authors bear all responsibility in case of violation of rights, and confirm that this dataset is
% open-sourced under the MIT license.

\end{document}